%% file: main.tex
\documentclass[11pt]{article}
\newlength{\defbaselineskip}
\usepackage{booktabs} 
\usepackage{amsmath}
\usepackage{amsthm}
\usepackage{amssymb}
\usepackage{algorithm, algorithmic}
\usepackage{multirow}
\usepackage{adjustbox}
\usepackage{nicefrac}
\usepackage{graphicx}
\usepackage{subcaption}

\usepackage{footnote}
\makesavenoteenv{tabular}
\makesavenoteenv{table}

\def\ba#1\ea{\begin{align*}#1\end{align*}}

\newcommand{\bi}{\begin{itemize}}
\newcommand{\ei}{\end{itemize}}
\newcommand{\be}{\begin{enumerate}}
\newcommand{\ee}{\end{enumerate}}
\newcommand{\bc}{\begin{center}}
\newcommand{\ec}{\end{center}}

\newcommand{\R}{\mathbb{R}}

\makeatletter
\newcommand\footnoteref[1]{\protected@xdef\@thefnmark{\ref{#1}}\@footnotemark}
\makeatother

\begin{document}
\title{Semi-Supervised Learning on Graphs Based on Local Label Distributions}

\author{Evgeniy Faerman\thanks{Ludwig-Maximilians-Universit\"at M\"unchen}
\and
Felix Borutta\footnotemark[1]
\and
Julian Busch\footnotemark[1]
\and
Matthias Schubert\footnotemark[1]
}

\maketitle
\begin{abstract}
Most approaches that tackle the problem of node classification consider nodes to be similar, if they have shared neighbors or are close to each other in the graph. Recent methods for attributed graphs additionally take attributes of neighboring nodes into account. We argue that the class labels of the neighbors bear important information and considering them helps to improve classification quality. Two nodes which are similar based on class labels in their neighborhood do not need to be close-by in the graph and may even belong to different connected components. 
In this work, we propose a novel approach for the semi-supervised node classification. Precisely, we propose a new node embedding  which is based on the  class labels in the local neighborhood of a node. We show that this is a different setting from attribute-based embeddings and thus, we propose a new method to learn label-based node embeddings which can mirror a variety of relations between the class labels of neighboring nodes. Our experimental evaluation demonstrates that our new methods can significantly improve the prediction quality on real world data sets.

\end{abstract}

\maketitle

\input{introduction}

\input{problem_setting}

\input{method}

\input{related_work}

\input{experiments}
\input{conclusion}

\bibliographystyle{IEEEtran}
\bibliography{sample-bibliography}

\end{document}

%% file: introduction.tex
\section{Introduction}
\label{sec:intro}

Graphs are the most general way to represent structured data. In general, a set of entities with some given pairwise relationships between them can be modeled as a graph $G=(V,E)$ with a corresponding node set $V$ and an edge set $E \subseteq V \times V$. Real-world examples of graph-structured data are abundant and include social networks, co-citation networks or biological networks.

In addition to the graph structure, further attribute information may be provided for the entities described by the graph nodes. In an attributed graph, each node $v_i \in V$ is associated with an attribute vector $f_i \in \R^d$. For instance, social network users might be enriched with personal information or documents in a co-citation network might be described by bag-of-words vectors. The increasing relevance of graph-structured data has been accompanied by an increased interest in learning algorithms which can leverage underlying graph structure to make accurate predictions for the modeled entities.

An important semi-supervised learning task on graphs is node classification, where each node $v_i \in V$ can be associated with a set of class labels (simply referred to as labels in the following) represented by a label vector $y_i \in \{0,1\}^l$ where $l$ is the number of possible labels. Given a set of already labeled nodes in a graph, the goal is to predict new likely labels for unlabeled nodes. The task is semi-supervised in the sense that connectivity information about the whole graph is available and at least some of the class labels are already known. In the case of attributed graphs, attributes of all nodes can additionally be used for prediction, including those of the unlabeled nodes in the graph. Important applications include recommendation in social networks, where the node labels represent user interests, or document classification in co-citation networks, where the node labels indicate associated fields of research.

Approaches for node classification on graphs may employ additional node attributes or operate on the graph structure alone. We will refer to these approaches as \emph{attribute-based} and \emph{connectivity-based} approaches, respectively.
Among the most successful connectivity-based methods are node embedding techniques \cite{perozzi2014deepwalk, grover2016node2vec, tang2015line, faerman2017lasagne, cao2015grarep, wang2016structural, ribeiro2017struc2vec, hamilton2017inductive, bojchevski2017deep, kipf2016variational}.
An underlying assumption of these techniques is that nodes which are closely connected in the graph, should have similar labels, which is commonly referred to as homophily \cite{mcpherson2001birds}.
Our method does not rely on the homophily assumption, but is still able to relate close-by nodes. Furthermore, unlike most node embedding techniques our new approach can be used to classify nodes unseen during training.
In the attribute-based setting the graph structure can be incorporated in different ways, for instance by using regularization \cite{zhu2003semi, zhou2004learning, belkin2006manifold, weston2012deep}, combining attributes with node embeddings \cite{yang2016revisiting}, or aggregating them over local neighborhoods \cite{DBLP:journals/corr/BrunaZSL13,DBLP:journals/corr/HenaffBL15, NIPS2016_6081, kipf2016semi, DBLP:journals/corr/AtwoodT15, DBLP:journals/corr/MontiBMRSB16, DBLP:journals/corr/LevieMBB17,simonovsky2017dynamic,velivckovic2017graph}. While regularization-based methods rely on the homophily assumption and most of them are not able to classify instances unseen during training, all other methods focus on node attributes.
In addition to connectivity and node attributes, the labels available during training further provide valuable information that is in general complementary to connectivity and attribute features, and are useful to improve classification.
In general learning tasks on independent and identically distributed (iid) data, labels indicate that an observation is sampled from a particular distribution. However, in a graph we have non-iid data and thus, the labels of connected objects allow for a novel use of label information which has not been exploited before for learning graph embeddings.

\begin{figure}
\centering
\includegraphics[width=0.65\columnwidth]{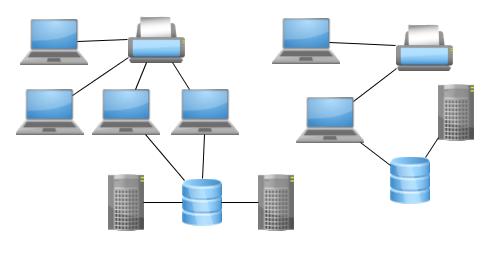}
\caption{Consider a communication network with nodes labeled according to their device type (user, server, database, printer). Assume the labels for the database and printer nodes in the right connected component are unknown while the remaining labels are provided. Further node attributes are not given. We can observe that the roles of printers and databases are clearly defined by the labels of their neighboring nodes, e.g., printers are not connected to server nodes.
Homophily-based methods would fail to classify these nodes correctly, since their labels differ from their neighbors. Further, connectivity alone does not explain the roles, since for instance the printer and database in the left part of the graph have the same degree and even their neighbors have the same degrees.
}
\label{fig:example}
\end{figure}

In this paper, we propose a \emph{label-based} approach to learn a node embedding which allows for more accurate node classifications. The main idea of our approach is that there often exists a correlation between the labels of a node and the distribution of labels in its local neighborhood. Thus, considering the local label distribution when computing a node embedding can exploit this correlation to improve the descriptiveness of the learned embedding.
In Figure \ref{fig:example}, we illustrate this for a typical case for which the label of a node is determined by the labels of neighboring nodes and not by node attributes or connectivity. As an additional example, the function of a protein can be expected to correlate strongly with the functions of interacting proteins.
As mentioned above, we assume that the labels of at least some of the neighboring nodes are known for each new node with unknown labels. In the majority of applications, this is realistic because new nodes usually connect to already known parts of the network. For instance, new papers usually cite established articles and new members of a social network will usually already know multiple friends in the network to connect to.

Though labels can be considered as another type of node attributes, there exists an important difference between labels and attributes which prevents
attribute-based embeddings to generalize well on label information. Though the attribute values of the predicted node are allowed to be used for learning
the embedding, using the node labels even in an transitive way leads to overfitting and  a bad generalization performance of the learned embedding.
We will discuss these issues in more detail in Section \ref{sec:method} and introduce a simple baseline method.
In our new method, we aggregate labels from relevant nodes directly and thus, we can completely exclude any influence of the nodes' own labels. In a first step, we determine the relevant neighbors of a given node based on \emph{Approximate Personalized PageRank (APPR)}. Since this might be an expensive task for large graphs, we use an adaption of the highly efficient algorithm from \cite{Shun:2016:PLG:2994509.2994522}. After determining the neighborhood, we compute the label distribution within the neighborhood and classify the node based on this novel representation.
We compare our new representation to state-of-the-art graph embeddings based on several benchmark datasets for node classification.

The remainder of the paper is structured as follows: After providing a formal problem definition for our approach in Section \ref{sec:problem_setting}, we introduce our new method in Section \ref{sec:method}, starting with a discussion on the possibility of incorporating label-based features into existing models in Section \ref{subsec:motivation}.
After a discussion of related work in Section \ref{sec:related_work}, the performance of our model is evaluated experimentally and compared to state-of-the-art methods in Section \ref{sec:experiments}. Finally, Section \ref{sec:conclusion} concludes the paper and proposes directions for future work.

%% file: problem_setting.tex
\section{Problem Setting}
\label{sec:problem_setting}

We consider (possibly directed) graphs $G=(V,E)$, with node set $V = \{v_1, \dots, v_n \}$ and edge set $E \subseteq V \times V$. A graph can be represented by an $n \times n$ adjacency matrix $A = (a_{ij})_{v_i,v_j \in V}$, where $a_{ij} \in \R$ denotes the weight of the edge $(v_i,v_j)$. In case of an unweighted graph,  $a_{i,j} = 1$ indicates the existence and  $a_{i,j} = 0$ the absence of an edge between $v_i$ and $v_j$. Furthermore, we do not allow self-links, i.e.,
 $a_{i,i} = 0$ for all nodes $v_i \in V$.
In an attributed graph, additional node attributes are provided in the form of an attribute vector $f_i \in \R^d$ for each node $v_i$. The attribute information for the whole graph can be represented by an $n \times d$ attribute matrix $F$, where the $i$th row of $F$ corresponds to $v_i$'s attribute vector $f_i$. Let us note that an important difference between attributes and labels is that attributes are usually known for all nodes, in particular those nodes without known labels.

Our problem setting is \emph{semi-supervised node classification}, where the node set $V$ is partitioned into a set of labeled nodes $L$ and unlabeled nodes $U$, such that $V = L \cup U$ and $L \cap U = \emptyset$. Thereby, each node $v_i \in V$ is associated with a label vector $y_i \in \{0, 1\}^l$, where $l$ is the number of possible labels and an entry one indicates the presence of the corresponding label for a certain node. The labels available for training can be represented by an $n \times l$ label matrix $Y_{\text{\textit{train}}}$, where the $i$'ths row of $Y_{\text{\textit{train}}}$ corresponds to the label vector $y_i$ of $v_i$ if $v_i \in L$. For unlabeled nodes, we assign constant zero vectors. The task is now to train a classifier using $A$, $Y_{\text{\textit{train}}}$ and possibly $F$ which accurately predicts $y_i$ for each $v_i \in U$. In \emph{multi-class} classification, each node is assigned to exactly one class, such that $y_i = e_j$ is the $j$'s unit vector, if $v_i$ is assigned to class $j$. \emph{Multi-label} classification denotes the general case, in which each node may be assigned to one or more classes and the goal is to predict all labels assigned to a particular node.

%% file: method.tex
\section{Semi-supervised learning on graphs based on local label distribution}
\label{sec:method}

\subsection{Labels as Attributes}
\label{subsec:motivation}

The main idea of our approach is to learn a more descriptive node representation by incorporating the known labels in the neighborhood of a node. 
In the following, we will show why existing methods  are not suitable to consider this information. Methods relying on neighborhood similarity \cite{perozzi2014deepwalk, grover2016node2vec, cao2015grarep, tang2015line,wang2016structural, faerman2017lasagne, bojchevski2017deep} learn representations in an unsupervised manner and thus, only rely on the topology of the graph and not on attributes or labels.
The \textit{Planetoid-T} model \cite{yang2016revisiting} considers labels by partly enforcing the similarity between members of the same class and therefore, nodes are related to each other based only on their own labels.

Graph Neural Networks \cite{scarselli2009graph,li2016gated} or Graph Convolution Networks (GCN) \cite{DBLP:journals/corr/BrunaZSL13,DBLP:journals/corr/HenaffBL15, NIPS2016_6081, kipf2016semi, DBLP:journals/corr/AtwoodT15, DBLP:journals/corr/MontiBMRSB16, DBLP:journals/corr/LevieMBB17,simonovsky2017dynamic,velivckovic2017graph} are special cases of a Message Passing Neural Network (MPNN) \cite{gilmer2017neural} which is a framework describing a family of neural network based models for attributed graphs.
All MPNN methods have in common that they use some differentiable function to iteratively compute messages for each node which are passed to all its neighbors. These messages build an input to a differentiable update function which computes new node representations $h$:
$$m_{v}^{t+1} = \sum_{w\in N(v)}M_{t}(h_{v}^t,h_{w}^t),$$
$$h_{v}^{t+1} = U_{t}(h_{v}^t, m_{v}^{t+1}).$$
Here $t$ denotes the current iteration, $h_{v}^t$ is the representation of node $v$ in iteration $t$ and vector $h_{v}^0$ corresponds to the input features of node $v$. $N(v)$ denotes the set of direct neighbors of node $v$, $M_{t}$ is the message and $U_{t}$ the set of update functions.
The obvious way to integrate the neighborhood label information into an MPNN-based prediction model is to include the label information into the messages directed to the neighbors in the first iteration.
However, even after removing self-links each node would receive information about its own labels already in the second iteration during the training stage. Thus, models learned on these representations overfit on the nodes' own labels and do not generalize well in the inference step where the node labels are unknown.
The same applies to directed graphs with cycles. Therefore, applying MPNN models to communicate neighboring labels is restricted to one iteration only. We use a corresponding model as a baseline for our experiments.

Note that this problem does not apply to label propagation algorithms \cite{zhou2004learning,zhu2003semi} since they do not have separate training and inference steps.
However, these methods have to be recomputed for predicting new objects which limits their usability.

\subsection{General Approach}

To present our method for semi-supervised learning on graphs using local label distributions we first outline an efficient algorithm for computing node neighborhoods based on \textit{Approximated Personalized PageRank} (APPR).
Afterwards, we describe how to create node representations based on the label distribution in the local neighborhood based on APPR. Finally, the node representations can be used as feature descriptors in arbitrary classification models.

The \textit{Personalized PageRank} (PPR) corresponds to the \textit{PageRank} algorithm \cite{page1999pagerank}, where the probabilities in the starting vector $s$ are biased towards some set of nodes. The result is the ``importance'' of all nodes in the graph from the viewpoint of the nodes in $s$.

The \textit{push} algorithm described in \cite{jeh2003scaling} and \cite{berkhin2006bookmark} is an efficient way to compute an approximation of the \textit{Personalized PageRank} (APPR) vector if the start distribution vector $s$ is sparse.
The idea behind the \textit{push} algorithm is only to consider a node in the local neighborhood if the probability to visit the node is significantly larger than the probability to visit any other node from the rest of the graph.
This leads to a sparse solution meaning that only relatively few nodes of the underlying graph are contained in the resulting APPR vector.

Algorithm \ref{alg:appr_ppr} describes the computation of APPR using a variant of the \textit{push} operation on lazy random walk transition matrices of undirected unweighted graphs. This algorithm was proposed in \cite{andersen2006local}, where APPR is used to partition graphs. We describe an adapted version from \cite{Shun:2016:PLG:2994509.2994522} which converges faster. 
The algorithm maintains two vectors: the solution vector $p$ and a residual vector $r$. The vector $p$  is the current approximation of the PPR vector and vector $r$ contains the approximation error or the not yet distributed probability mass. $p(u)$ and $r(u)$ are the entries in vectors $p$ and $r$ corresponding to node $u$, $d(u)$ is the degree of node $u$. In each iteration the algorithm selects a node with sufficient probability mass in vector $r$. This probability mass is spread  between the node entry in $p$ and the entries of its direct neighbors in $r$.
In each step, the exact PPR is the linear combination of the current solution vector $p$ and the PPR solution for $r$, i.e., $pr(s)=p+pr(r)$. This procedure can also be trivially adapted to directed graphs and graphs with weighted edges.
Moreover, the algorithm requires two parameters: $\alpha$, which is the teleportation parameter and determines the level of locality for each node neighborhood; and $\epsilon$, which is an approximation threshold and controls the approximation quality and the runtime. In fact, the \textit{push} algorithm performs updates as long as there is one node for which at least $\epsilon \frac{1-\alpha}{1+\alpha}$ probabilty mass is moved towards each of its neighbors. The complexity of the procedure is $\frac{1}{\alpha \cdot \epsilon}$.
\begin{algorithm}[t]
	\algsetup{linenosize=\fontsize{7}{8.4}}
	\fontsize{7}{8.4}
	\selectfont
	\caption{ApproximatePPR}\label{alg:appr_ppr}
	\begin{algorithmic}[1]
		\renewcommand{\algorithmicrequire}{\textbf{Input:}}
		\renewcommand{\algorithmicensure}{\textbf{Output:}}
		\REQUIRE Starting vector $s$, Teleportation probability $\alpha$, Approximation threshold $\epsilon$
		\ENSURE APPR vector $p$
		\STATE $p=\vec{0}$, $r = s$
		\WHILE {$r(u) \geq \epsilon d(u)$  for some vertex $u$}
			\STATE pick any $u$ where $r(u) \geq \epsilon d(u)$
			\STATE push(u)
		\ENDWHILE
		\RETURN $p$
		\medskip
	\end{algorithmic}
	\vspace{-1ex}
\end{algorithm}

\begin{algorithm}[t]
	\algsetup{linenosize=\fontsize{7}{8.4}}
	\fontsize{7}{8.4}
	\selectfont
	\caption{push}\label{alg:push}
	\begin{algorithmic}[1]

		\STATE $p(u) = p(u)+(2\alpha/(1+\alpha))r(u)$
		\FOR {$v$ with $(u,v)\in E$}
			\STATE $r(v) = r(v) + ((1-\alpha)/(1+\alpha))r(u)/d(u)$ \label{alg:push:line_neigh_update}
		\ENDFOR
		\STATE $r(u) = 0$
		\medskip
	\end{algorithmic}
	\vspace{-2ex}
\end{algorithm}

\subsubsection{Local Label Distribution}
In our approach we first compute the APPR vector for each node. Before APPR is computed for node $v$, the corresponding entry $s(v)$ in starting vector $s$ is set to one and all other entries to zero.
Therefore, the APPR vector of $v$ describes the importance of local neighbors only from its point of view.

In the APPR result matrix $APPR$, each row corresponds to the APPR vector of the corresponding node.
The local label distribution representation $X\in \mathbb{R}^{n \times l}$ is computed 
by multiplying $APPR$ with the label matrix $\mathbf{Y}_{\text{\textit{train}}}$. The diagonal of $APPR$ is set to zero beforehand to exclude information about the own labels. Therefore, the entry $X_{v\:y_{j}}$ can be interpreted as the probability that a random walk starting from node $v$ stops at a neighbor with label $y_{j}$.

The local label distribution can be used as a node embedding vector which can be passed into an arbitrary classification algorithm. In our experiments, we
employ a multi-layer perceptron.

%% file: related_work.tex
\section{Related Work}
\label{sec:related_work}

Numerous approaches for semi-supervised learning on graphs have been proposed in recent years. This can be sorted into two main categories, unsupervised node embedding techniques and semi-supervised techniques.

\subsection{Unsupervised Node Embedding}
\label{subsec:semi_graphs}

A strong focus of recent developments related to learning from structural relationships has been placed on learning \emph{node embeddings}, where a latent vector representation is learned for each node, reflecting its connectivity in the underlying graph. The learned node embeddings can be used as an input to a subsequent down-stream task, such as node classification.
Random walk based methods \cite{perozzi2014deepwalk,grover2016node2vec} sample a number of random walks from the graph and nodes are related if they have common neighbors. \emph{LINE} \cite{tang2015line} is another variant, which considers direct first- and second-order proximities instead of random walks. \emph{Graph2Gauss} \cite{bojchevski2017deep} learns similarity to hop neighborhoods and embeds each node as a Gaussian distribution to allow for uncertainty in the representation. \emph{GECS} \cite{al2016gecs} uses connections subgraphs to determine appropriate node neighborhood. More closely related to our approach, \emph{LASAGNE} \cite{faerman2017lasagne} relies on APPR to determine relevant context nodes. Other works perform matrix factorization. For instance, \emph{GraRep} \cite{cao2015grarep} factorizes a sequence of $k$-step log-probability matrices with SVD and concatenates the resulting low-dimensional node representations to form the final representations. Abu-El-Haija et al. propose matrix factorization of random-walk occurrence matrix with different approaches to determine context window size distribution \cite{abu2017watch}. \emph{SDNE} \cite{wang2016structural} uses a multi-layer auto-encoder model to capture non-linear structures based on direct first- and second-order proximities. Authors of \cite{chamberlain2017neural} propose embeddings in hyperbolic space. \emph{HARP} \cite{chen2017harp} addresses the local minima problem and introduce an iterative scheme for learning of node representations which can be used with different embedding learning methods. An input graph is coarsened on different levels and node representations are learned starting with the coarsest graph and learned embeddings are provided as initializations for the embeddings of subsequent finer graphs.
While the above methods rely on the homophily assumption, \emph{struc2vec} \cite{ribeiro2017struc2vec} aims at learning representations which relate structurally similar nodes instead of nodes which are close in the graph. It does so by using degree sequences in neighborhoods of different sizes.
All of the above approaches are \emph{transductive} in the sense that labels can only be predicted for unlabeled nodes observed already at training time. The \emph{GraphSAGE} \cite{hamilton2017inductive} framework introduces \emph{inductive} node embeddings. The basic idea is to learn an embedding function by sampling and aggregating node attributes in local neighborhoods. The embedding function can further be learned with a supervised loss function.  Inductive models are also obtained by considering node attributes. \emph{Variational Graph Auto-Encoders} \cite{kipf2016variational} learn node representations using a variational auto-encoder, where the encoder is a two-layer GCN. The model can be applied to attributed and non-attributed graphs.

\subsection{Semi-Supervised Learning on Graphs}

Compared to separately optimizing steps in a semi-supervised learning pipeline, as is the case for semi-supervised learning with pre-trained node embeddings, end-to-end training usually leads to better performance on the supervised learning objective.

One important direction is \emph{Laplacian Regularization}, where the prediction loss is augmented with an unsupervised loss function based on the graph's Laplacian matrix, encoding the homophily assumption that close-by nodes should have the same label. Related approaches include \emph{Manifold Regularization} \cite{belkin2006manifold}, a kernel-based method, and \emph{Deep Semi-Supervised Embedding} \cite{weston2012deep} which incorporates node embeddings by augmenting neural network models with an embedding layer. Both of these methods generalize to attributed graphs. \emph{Label Propagation} \cite{zhu2003semi, zhou2004learning} is more closely related to our approach. The main idea is to propagate the observed labels through the graph via the random walk transition matrix. Similarly, \emph{Collective Classification} \cite{sen2008collective} starts with the observed labels and iteratively classifies nodes based on previously inferred labels in the local neighborhood based on a majority vote. However, none of these two methods considers additional node attributes and no actual learning is involved. Similarly to collective classification, the more recent methods \emph{Structural Neighborhood Based Classification} \cite{nandanwar2016structural} and \emph{Weighted-Vote Geometric Neighbor Classification} \cite{ye2017learning} also classify nodes based on labels in local neighborhoods. They achieve this by learning a model which predicts node labels from a feature vector describing the local $k$-neighborhood. Both methods assume an unattributed graph.

Instead of imposing regularization, \emph{Planetoid} \cite{yang2016revisiting} combines the prediction loss with node embeddings by training a joint model which predicts class labels as well as graph context for a given node. The graph context sampled from random walks as well as the set of nodes with shared labels. This allows Planetoid to relate nodes with similar labels even if they are not close in the graph. Thus, Planetoid does not rely on a strong homophily assumption. In addition to a connectivity-based variant, \emph{Planetoid-G}, the authors propose two further architectures, which incorporate node attributes. The transductive variant \emph{Planetoid-T} starts with pre-trained embeddings and alternately optimizes the prediction and embedding loss functions. The inductive variant \emph{Planetoid-I} on the other hand predicts the graph context from the node features instead.

Another important direction which has recently gained increasing attention is concerned with generalizing deep neural network architectures to graph-structured domains.
As the general approach consists of incorporating graph structure into supervised learning, these models assume an attributed graph. However, they can naturally be applied to non-attributed graphs by using the identity matrix as the attribute matrix. The vast majority of neural network based models for semi-supervised learning on graphs can be described within a message-passing framework. In a \emph{Message Passing Neural Network (MPNN)} \cite{gilmer2017neural}, each node has a hidden state which is updated iteratively during training. The initial hidden state of a node corresponds to it's attribute vector. In a first step, messages from $v_i$'s neighborhood are received and aggregated, where a message from neighbor $v_j$ depends on $v_i$'s and $v_j$'s hidden states. In a second step, $v_i$'s state is updated by combining it with the aggregated messages.
An important special case are \emph{Graph Convolution Networks} \cite{DBLP:journals/corr/BrunaZSL13, DBLP:journals/corr/HenaffBL15, NIPS2016_6081, kipf2016semi,  DBLP:journals/corr/MontiBMRSB16, DBLP:journals/corr/LevieMBB17, velivckovic2017graph}
which aggregate node attributes over local neighborhoods with spatially localized filters, similar to classical convolutional networks on images \cite{lecun1989backpropagation}. The \emph{ChebNet} \cite{NIPS2016_6081} aggregates messages from neighbors analogously to the eigenvectors of the graph's Laplacian matrix. The update function ignores the previous state and applies a non-linear activation. The resulting filters are $k$-localized. The \emph{GCN} \cite{kipf2016semi} is a simplification of the ChebNet, which only considers one-hop neighbors. Messages are aggregated according to a normalized adjacency matrix. In the update phase, the aggregated messages are multiplied with a learned filter matrix with a ReLU activation. For graph convolution networks, the number of message passing iterations corresponds to the number of layers.

%% file: experiments.tex
\begin{figure*}[t]
    \centering
    \begin{subfigure}[t]{0.45\columnwidth}
        \includegraphics[width=\textwidth]{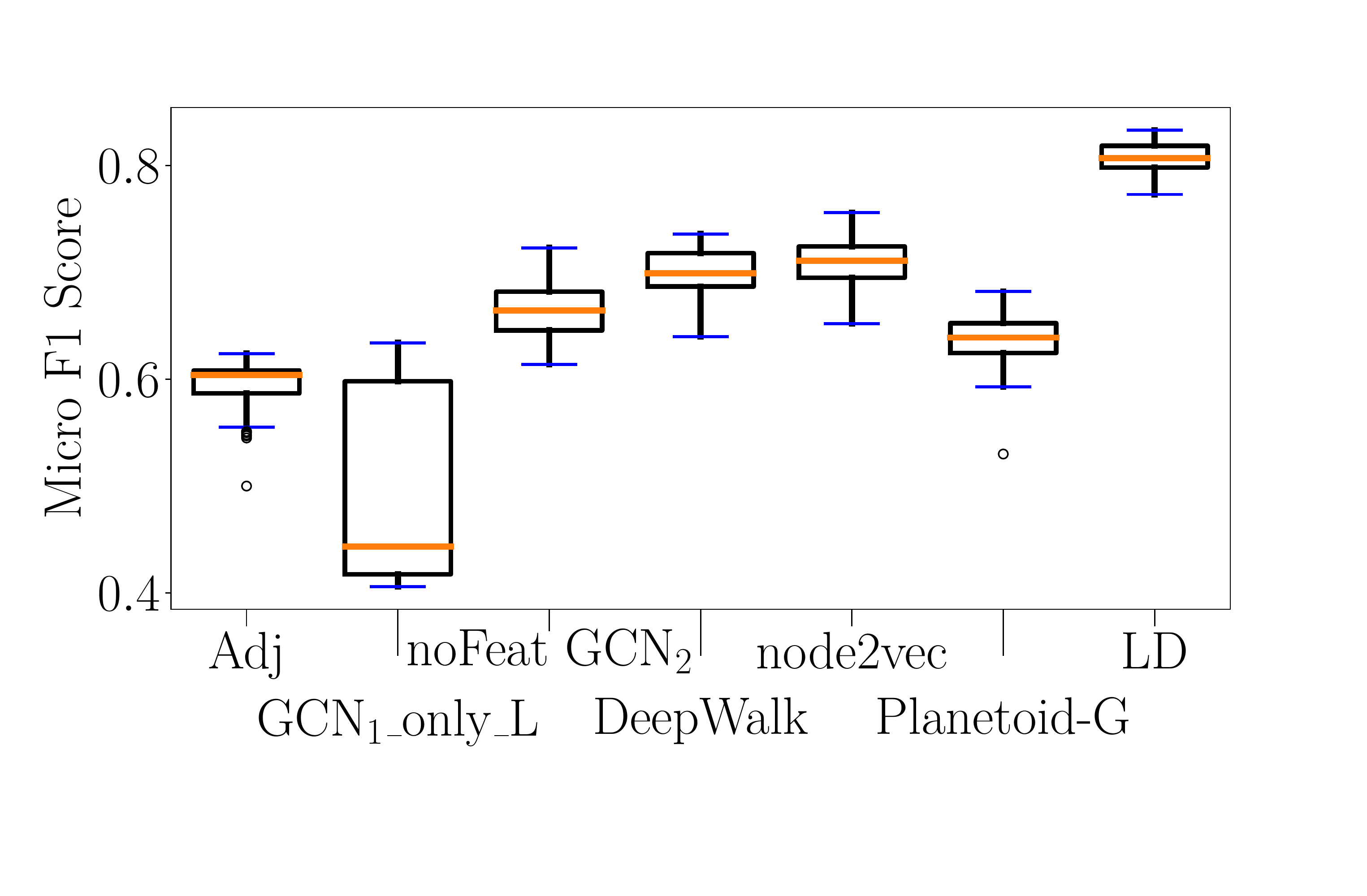}
        \vspace{-7ex}
        \caption{Micro F$_1$ scores for \textit{Cora}.}
        \label{fig:cora}
    \end{subfigure}
    \begin{subfigure}[t]{0.45\columnwidth}
        \includegraphics[width=\textwidth]{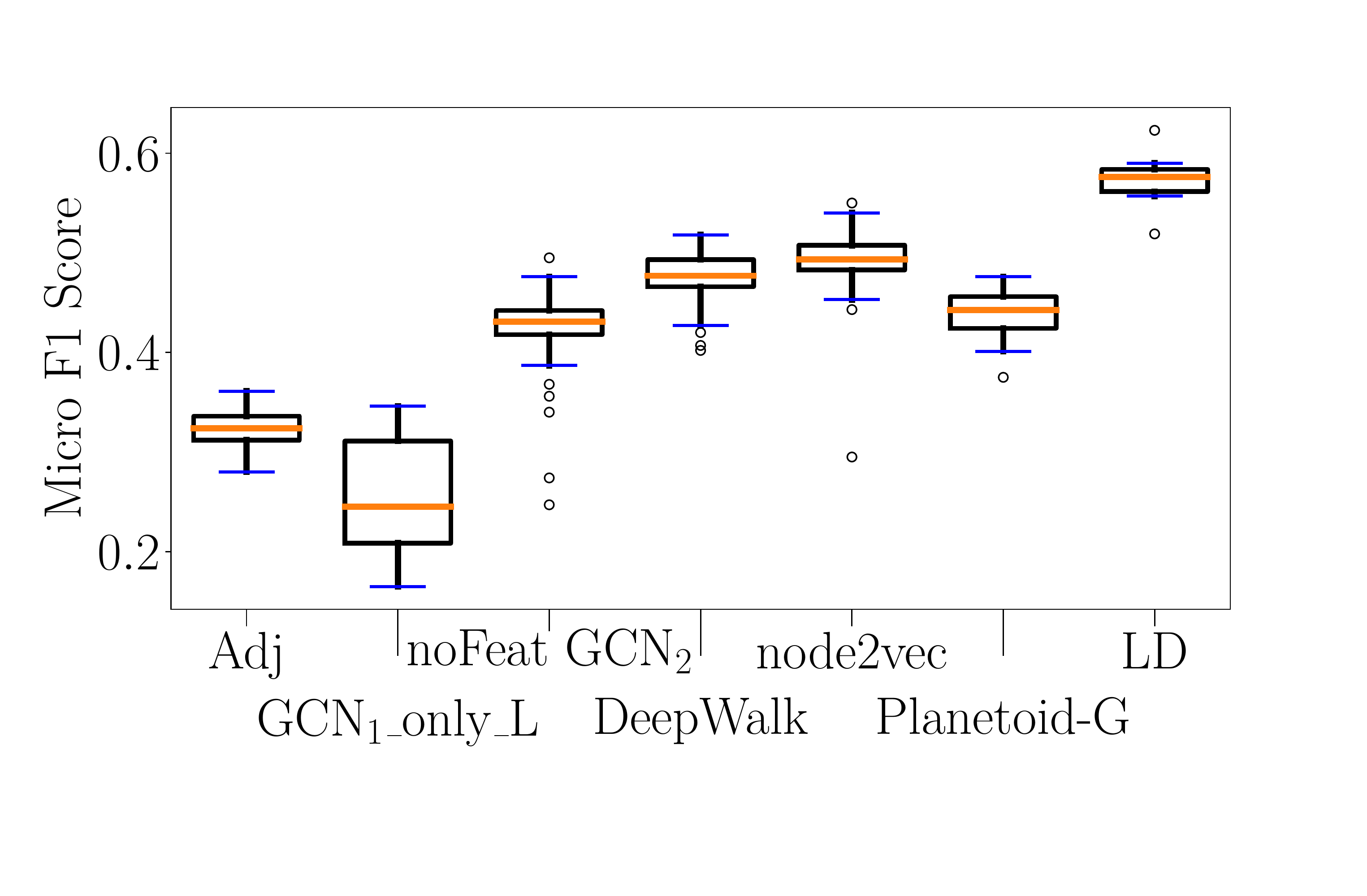}
        \vspace{-7ex}
        \caption{Micro F$_1$ scores for \textit{CiteSeer}.}
        \label{fig:citeseer}
    \end{subfigure}
    \begin{subfigure}[t]{0.45\columnwidth}
        \includegraphics[width=\textwidth]{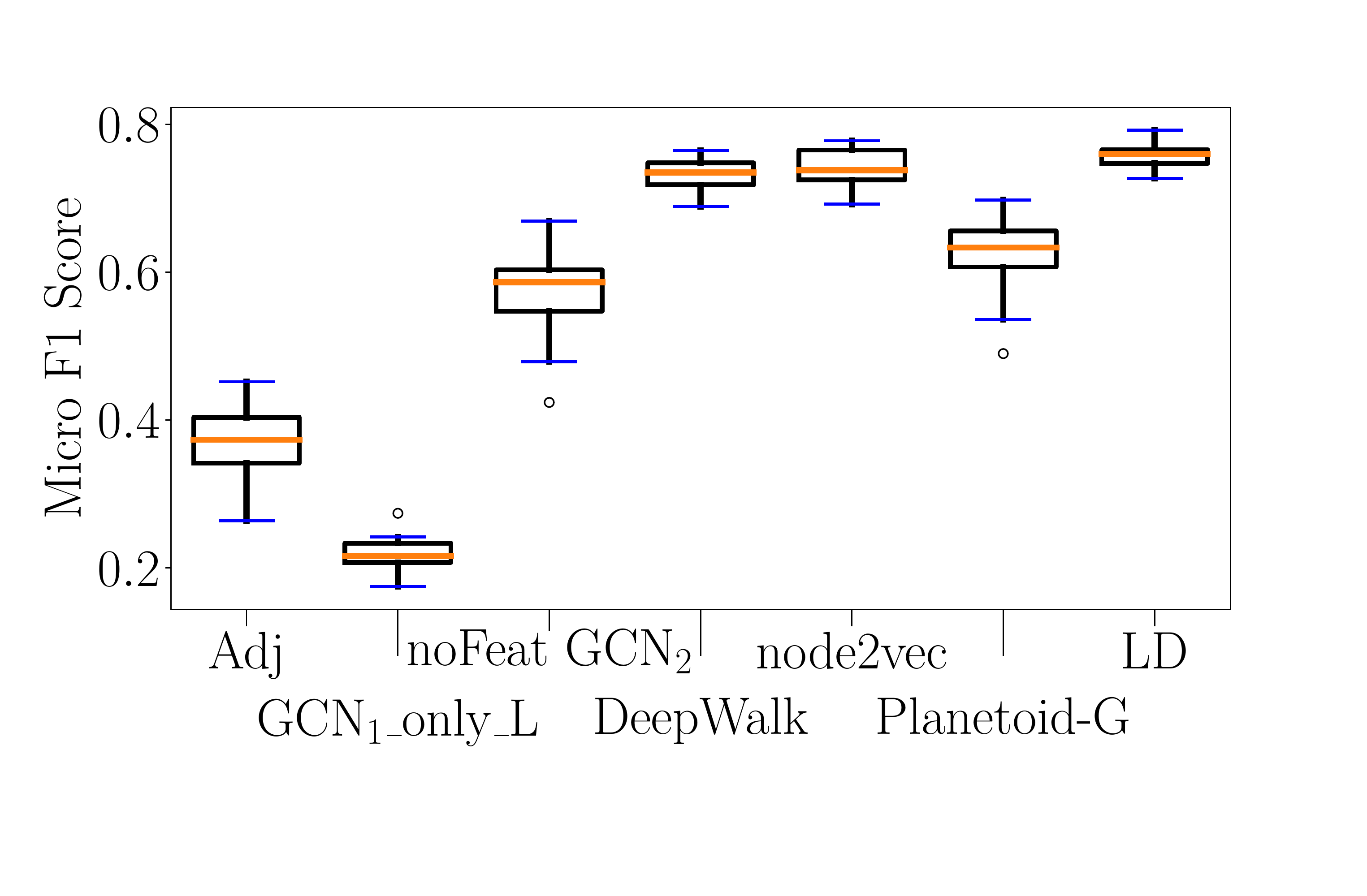}
        \vspace{-7ex}
        \caption{Micro F$_1$ scores for \textit{Pubmed}.}
        \label{fig:pubmed}
    \end{subfigure}
    \caption{Micro F$_1$ scores for the three benchmark data sets.}\label{fig:acc}
\end{figure*}

\section{Evaluation}\label{sec:experiments}
We evaluate our approach by performing node-label prediction and compare the quality in terms of micro F$_1$ score 
for multiclass prediction tasks, respectively micro F$_1$ and macro F$_1$ scores for multilabel prediction tasks, against state-of-the-art methods.

For both tasks, we compare our model against the following approaches:
\begin{itemize}
\item \textit{Adj}: a baseline approach which learns node embeddings only based on the information contained in the adjacency matrix
\item \textit{GCN$_1$\_only\_L}: a GCN which applies convolution on label matrix $\mathbf{Y}$. We use one convolution layer on the adjacency matrix without self-links, followed by a dense output layer\footnote{\label{foot:why_one_layer_gcn}See \ref{subsec:motivation} for the explanation why only one convolution layer makes sense}
\item \textit{noFeat GCN$_2$}: the standard 2-layer GCN as published by Kipf et al. \cite{kipf2016semi} without using the node attributes
\item \textit{DeepWalk}: the DeepWalk model as proposed in \cite{perozzi2014deepwalk}
\item \textit{node2vec}: the node2vec model as proposed in \cite{grover2016node2vec}
\item \textit{Planetoid-G}: the Planetoid variant which does not use attribute information \cite{yang2016revisiting}
\footnote{Unless stated differently we use for all competitors the parameter settings as suggested by the corresponding authors. Except for minor adaptations, e.g., to include label information in the one layer GCN models or to make the Planetoid models applicable for multilabel prediciton tasks, we use the original implementations as published by the correpsonding authors.}
\end{itemize}

Our model is denoted as \textit{LD} (short for \underline{L}abel \underline{D}istribution). For these experiments we train a simple feed-forward neural network which takes the label distribution based representations as input and retrieves class probabilities as output.

Note that we omit the comparison to label propagation \cite{zhu2003semi} since Yang et al. already showed that the \textit{Planetoid} model outperforms this approach \cite{yang2016revisiting}.

\vspace{-1ex}
\subsection{Multiclass Prediciton}

\subsubsection{Experimental Setup}\label{sec:setup}
For the multiclass label prediciton task we use the following three text classification benchmark graph datasets \cite{sen2008collective, namata2012query}:
\begin{itemize}
\item \textsc{Cora}. The Cora dataset contains 2'708 publications from seven categories in the area of ML. The citation graph consists of 2'708 nodes, 5'278 edges, 1'433 attributes and 7 classes.
\item \textsc{CiteSeer}. The CiteSeer dataset contains 3'264 publications from six categories in the area of CS. The citation graph consists of 3'264 nodes, 4'536 edges, 3'703 attributes and 6 classes.
\item \textsc{Pubmed}. The Pubmed dataset contains 19'717 publications which are related to diabetes and categorized into 3 classes. The citation graph consists of 19'717 nodes, 44'324 edges, 500 attributes and 3 classes.
\end{itemize}
\noindent For each graph, documents are denoted as nodes and undirected links between documents represent citation relationships. If node attributes are applied, bag-of-words representations are used as attribute vectors for each document.

We split the data as suggested in \cite{yang2016revisiting}, i.e., for labeled data our training sets contain 20 randomly selected instances per class, the test sets consist of 1'000 instances, and the validation sets contain 500 instances for each method. The remaining instances are used as unlabeled data. For comparison we use the prediction micro F$_1$ scores which we collected over 10 different data splits.

Since the numbers of iterations for sampling the graph contexts and the label contexts for \textit{Planetoid} are suggested only for the \textit{CiteSeer} data set, we adapted these values relative to the number of nodes for each graph. For \textit{node2vec}, we perform grid searches over the hyperparameters $p$ and $q$ with $p,q \in \lbrace 0.25, 0.5, 1.0, 2.0, 4.0\rbrace$ and use window size 10 as proposed by the authors. 
For all models except \textit{Planetoid} unless otherwise noted, we use one hidden layer with 16 neurons and regularization, learning rate and training procedure as in \cite{kipf2016semi}.  Considering our model, we use $\alpha \in \lbrace 0.1, 0.2, \ldots, 0.9\rbrace$ as values for the teleportation parameter and $\epsilon = 1e^{-5}$ as approximation threshold to compute the APPR vectors for each node.

We present results computed on the test sets for the best performing hyperparameters. The best performing hyperparameters for all models are determined by using the validation sets.

\subsubsection{Results}\label{sec:results}
Figure~\ref{fig:acc} shows boxplots depicting the micro F$_1$ scores we achieved for the multiclass prediction task for each considered model on the \textit{Cora}, \textit{CiteSeer} and \textit{Pubmed} networks.

The baseline approach \textit{GCN$_1$\_only\_L}, i.e., the one layer GCN model which only uses the label distributions of the neighboring nodes to predict a node's label, shows worst results among the considered models. However, these scores are still promising that the labels may improve the task of learning ``good'' representations. The baseline method which considers the corresponding rows of the adjacency matrix as node representations, i.e., \textit{Adj}, achieves slightly better results for all three datasets. For the \textit{GCN} and \textit{Planetoid} models that do not make recourse to attribute information, i.e., \textit{noFeat GCN$_2$}, resp. \textit{Planetoid-G}, the retrieved micro F$_1$ values are slightly lower than the ones achieved by \textit{DeepWalk} and \textit{node2vec}.
Our model improve the results produced by \textit{node2vec}, which means that the label distributions are indeed a useful source of information, although the baseline \textit{GCN$_1$\_only\_L} shows, especially for \textit{Pubmed}, rather poor results. This may be reasoned by the fact that this model only considers the label distribution of a very local neighborhood (in fact one hop neighbors). However, collecting the label distribution from a more spacious neighborhood gives a significant boost in terms of prediction accuracy. Indeed the best results for the \textit{LD} approach are reached for $\alpha = 0.1$, which corresponds to a rather spacious neighborhood exploration.

\vspace{-1ex}
\subsection{Multilabel Classification}
\begin{figure}
    \centering
    \begin{subfigure}[b]{0.45\columnwidth}
        \includegraphics[width=\textwidth]{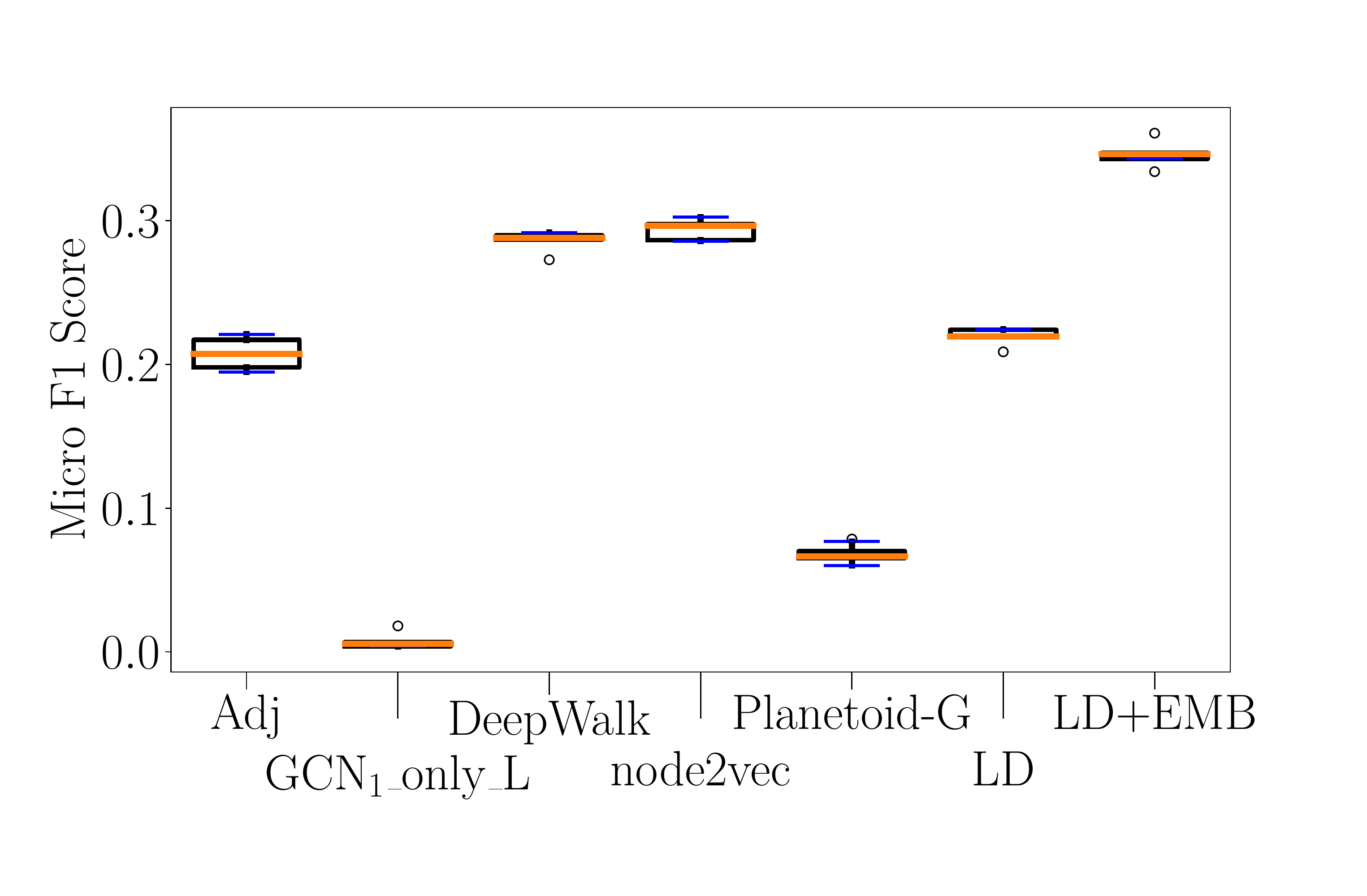}
        \vspace{-6ex}
        \caption{Micro F$_1$ scores for \textit{BlogCatalog}.}
        \vspace{-0ex}
        \label{fig:blogcatalog_micro}
    \end{subfigure}
    \begin{subfigure}[b]{0.45\columnwidth}
        \includegraphics[width=\textwidth]{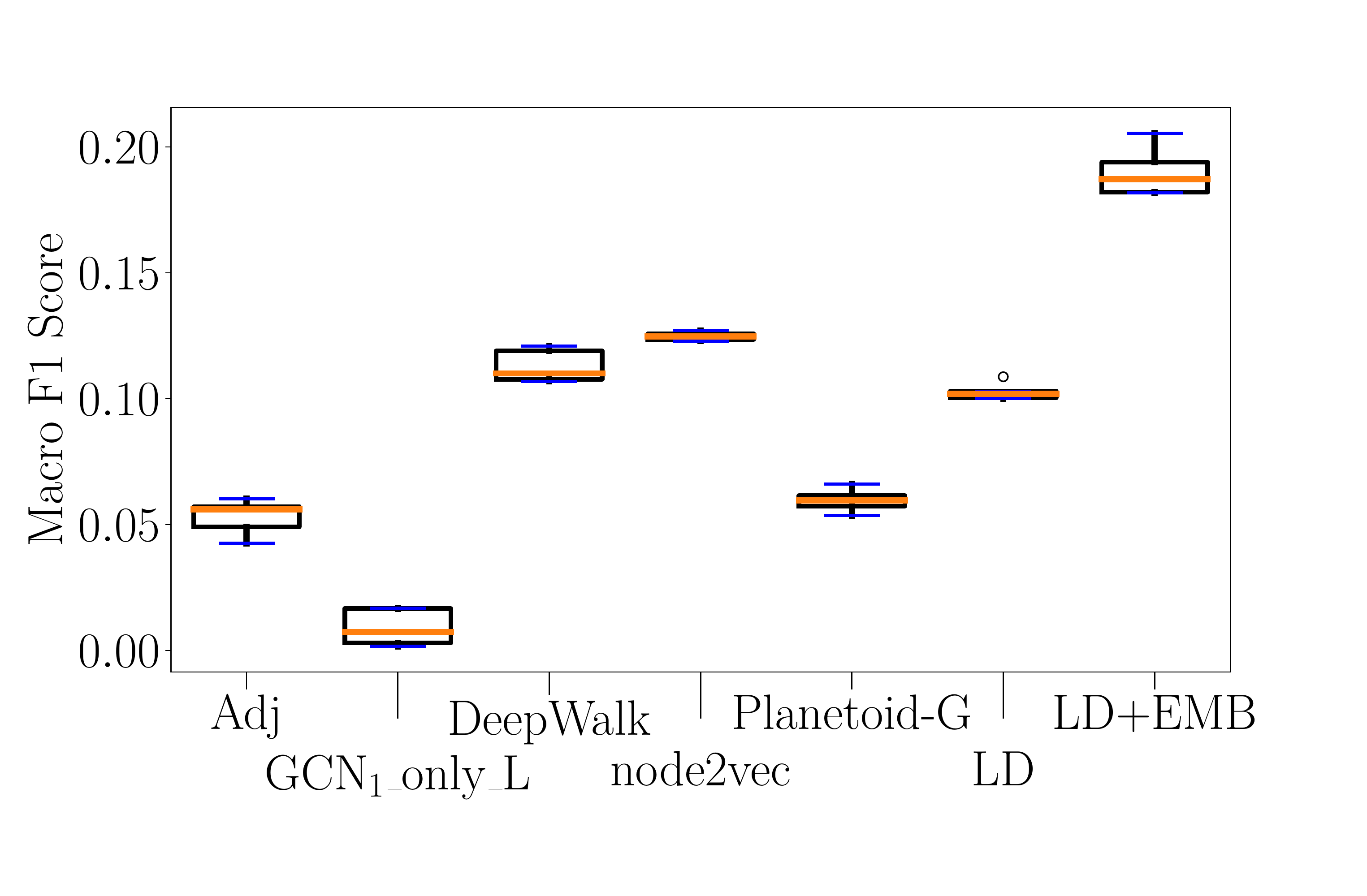}
        \vspace{-6ex}
        \caption{Macro F$_1$ scores for \textit{BlogCatalog}.}
        \label{fig:blogcatalog_macro}
    \end{subfigure}
    \caption{Micro F$_1$ and macro F$_1$ for \textit{BlogCatalog}.}\label{fig:blogcatalog}
\end{figure}

\subsubsection{Experimental Setup}
We also perform multilabel node classifications on the following two multilabel networks:

\begin{itemize}
\item \textsc{BlogCatalog} \cite{tang2009relational}. This is a social network graph where each of the 10,312 nodes corresponds to a user and the 333,983 edges represent the friendship relationships between bloggers. 39 different interest groups provide the labels.
\item \textsc{IMDb Germany}. This dataset is taken from \cite{faerman2017lasagne}. It consists of 32,732 nodes, 1,175,364 edges and 27 labels. Each node represents an actor/actress who played in a German movie. Edges connect actors/actresses that were in a cast together and the node labels represent the genres that the corresponding actor/actress played.
\end{itemize}
Since the fraction of positive instances is relatively small for most of the classes, we use weighted cross-entropy as loss function. Therefore, the loss caused by erroneously classified positive instances is weighted higher. We use weight 10 in all our experiments. For the same reason we report micro F$_1$ and macro F$_1$ score metrics to measure the quality of the considered methods.
We compare our model to the featureless models that we already used for the multiclass experiments \footnote{To adapt the \textit{Planetoid-G} implementation for multilabel classification, we use a \textit{sigmoid} activation function at the output layer and also slightly changed the embedding learning step. Entities that are used as context and have the same labels as the node itself are sampled from all classes to which the node belongs to.}.

\begin{figure}
    \centering
    \begin{subfigure}[b]{0.45\columnwidth}
        \includegraphics[width=\textwidth]{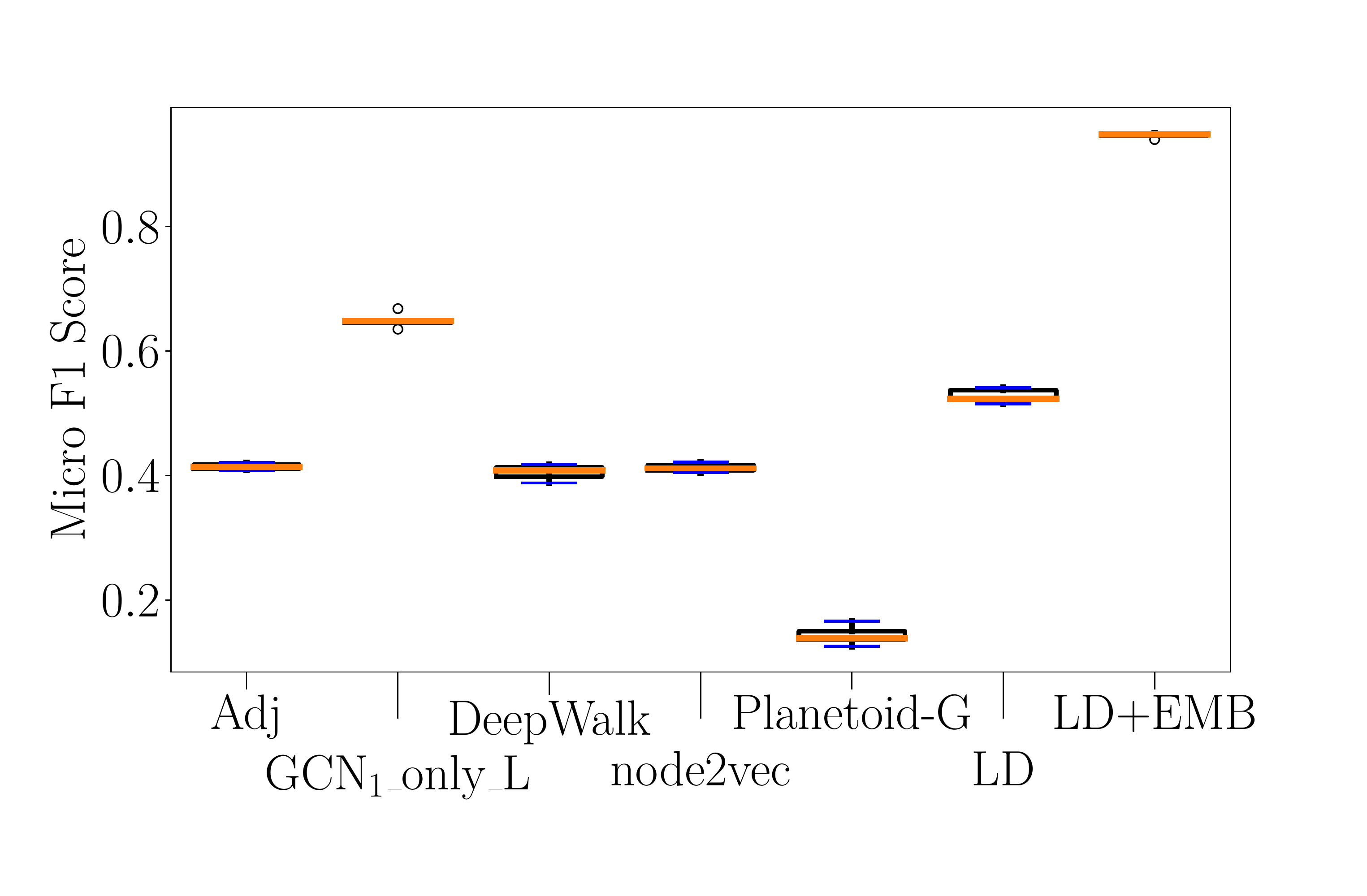}
        \vspace{-6ex}
        \caption{Micro F$_1$ scores for \textit{IMDb Germany}.}
        \label{fig:imdb_micro}
    \end{subfigure}
    \begin{subfigure}[b]{0.45\columnwidth}
        \includegraphics[width=\textwidth]{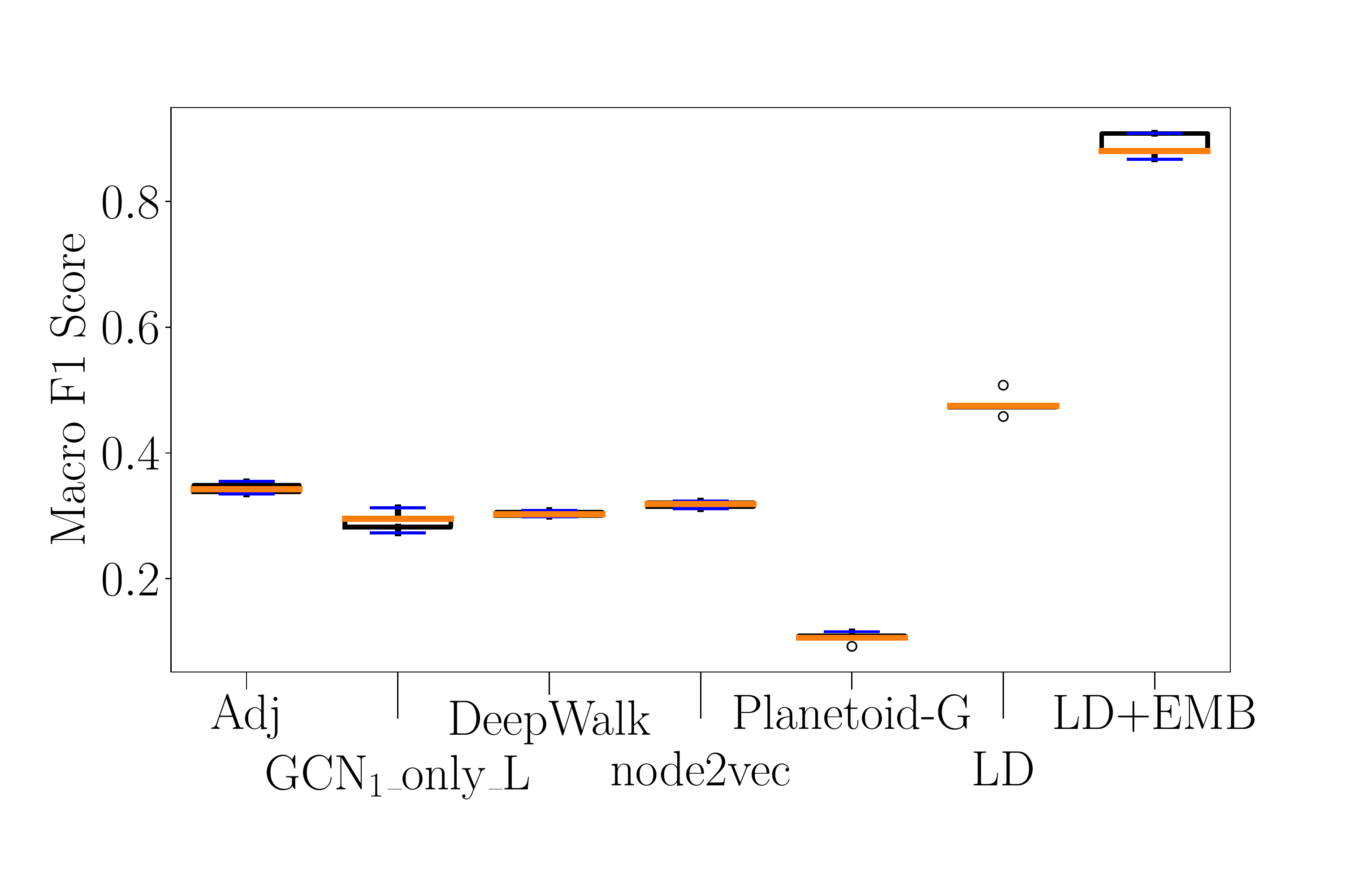}
        \vspace{-6ex}
        \caption{Macro F$_1$ scores for \textit{IMDb Germany}.}
        \label{fig:imdb_macro}
    \end{subfigure}
    \caption{Micro F$_1$ and macro F$_1$ for \textit{IMDb Germany}.}\label{fig:imdb}
\end{figure}

We split the data into training, validation and test set so that 70\% of all nodes were used for training, 10\% for validation and 20\% of the data were used to test the model. Note that we could not use stratified sampling splits for these experiments since we optimize for all classes simultaneously instead of using one-vs-rest classifiers \footnote{That is why our results for \textit{node2vec} and \textit{DeepWalk} on the BlogCatalog network are slightly worse than reported in \cite{grover2016node2vec}}. The hyperparameter setting is as described above. For this set of experiments we ran each model, except for \textit{Planetoid-G}, 10 times on five different data splits. Due to the long runtime of \textit{Planetoid-G} we trained this model only three times on two data splits.

\begin{figure*}[htb!]
    \centering
    \begin{subfigure}[t]{0.45\columnwidth}
        \includegraphics[width=\textwidth]{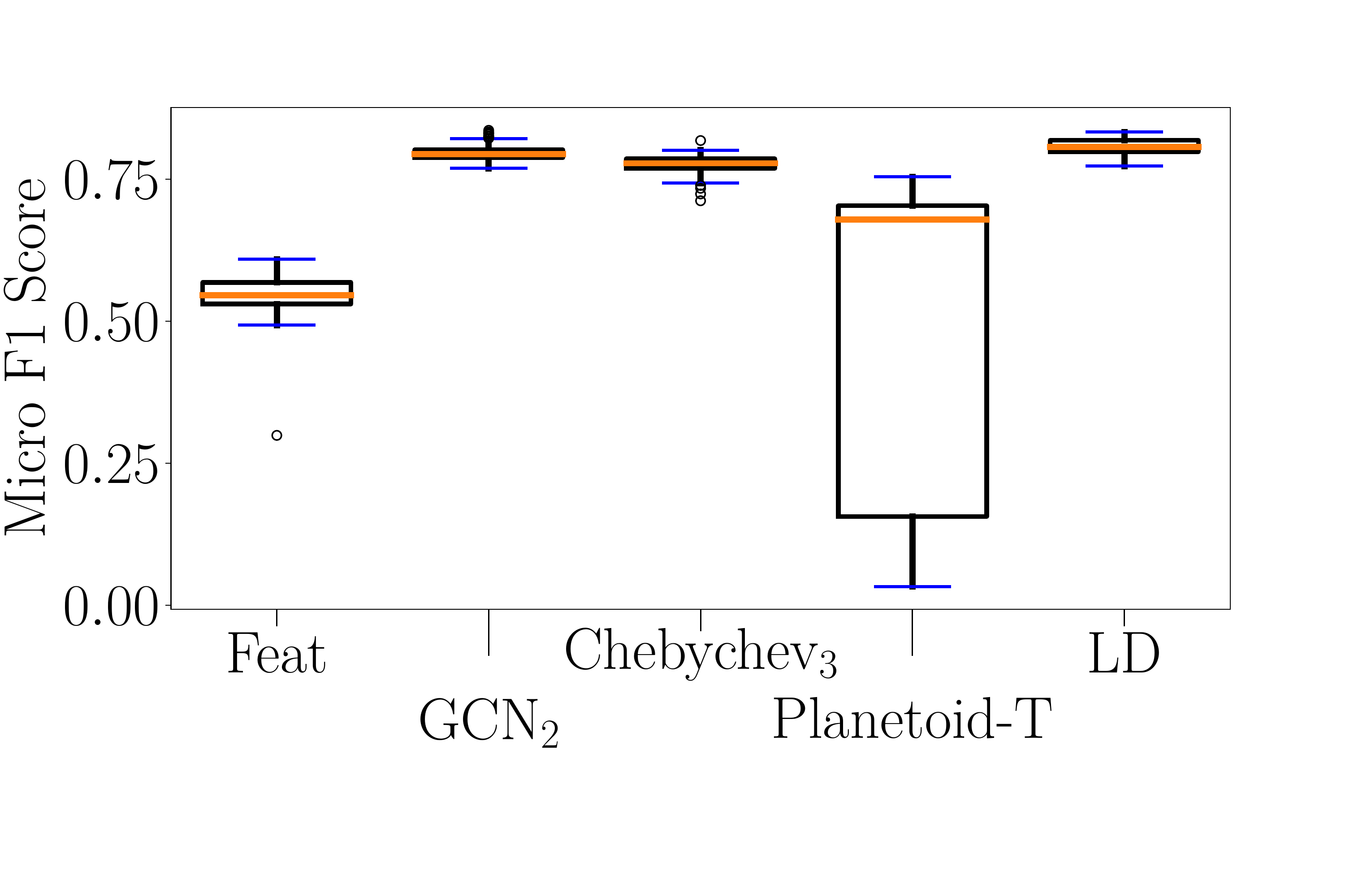}
        \vspace{-7ex}
        \caption{Micro F$_1$ scores for \textit{Cora}.}
        \label{fig:cora_feat}
    \end{subfigure}
    \begin{subfigure}[t]{0.45\columnwidth}
        \includegraphics[width=\textwidth]{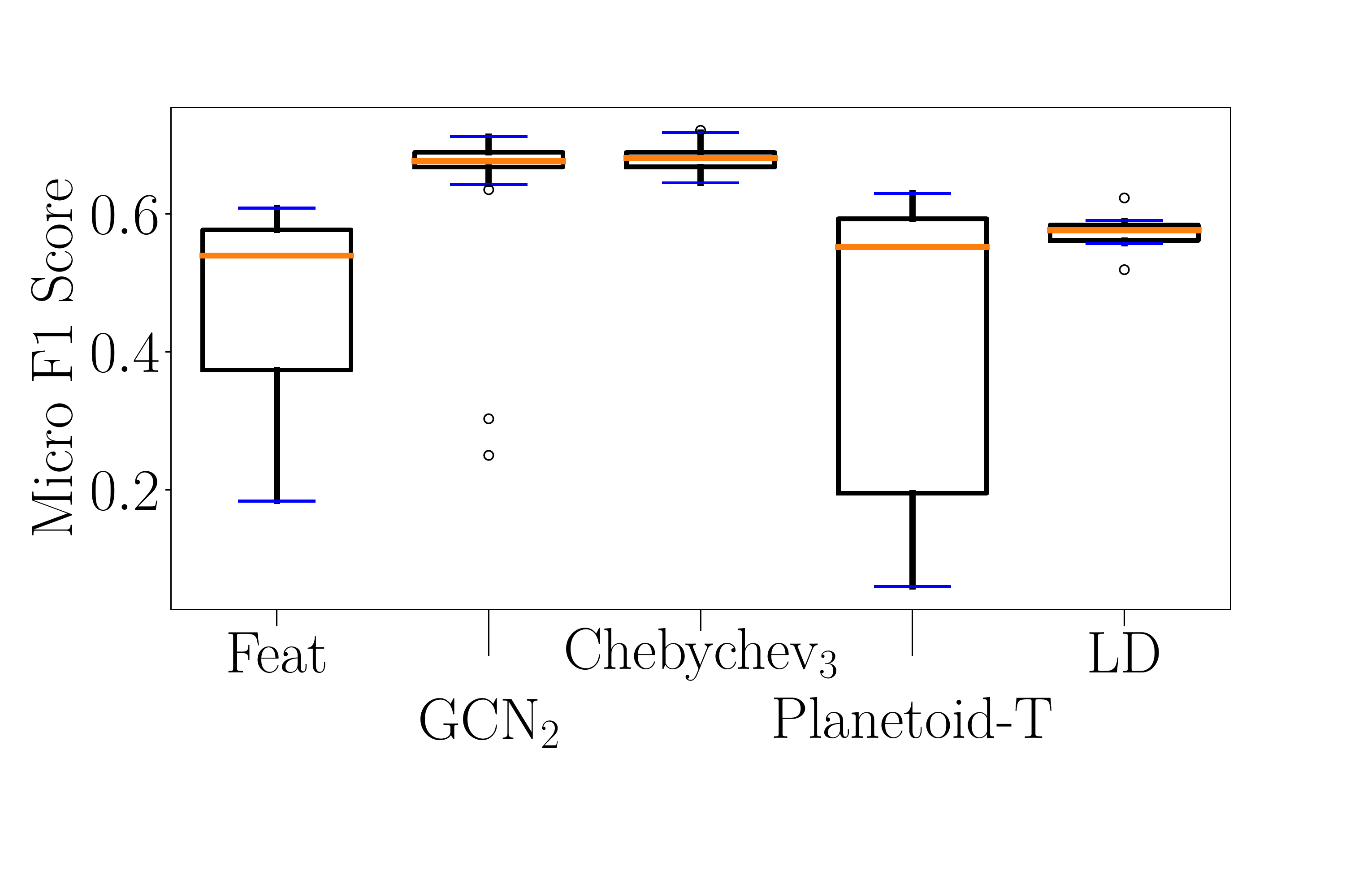}
        \vspace{-7ex}
        \caption{Micro F$_1$ scores for \textit{CiteSeer}.}
        \label{fig:citeseer_feat}
    \end{subfigure}
    \begin{subfigure}[t]{0.45\columnwidth}
        \includegraphics[width=\textwidth]{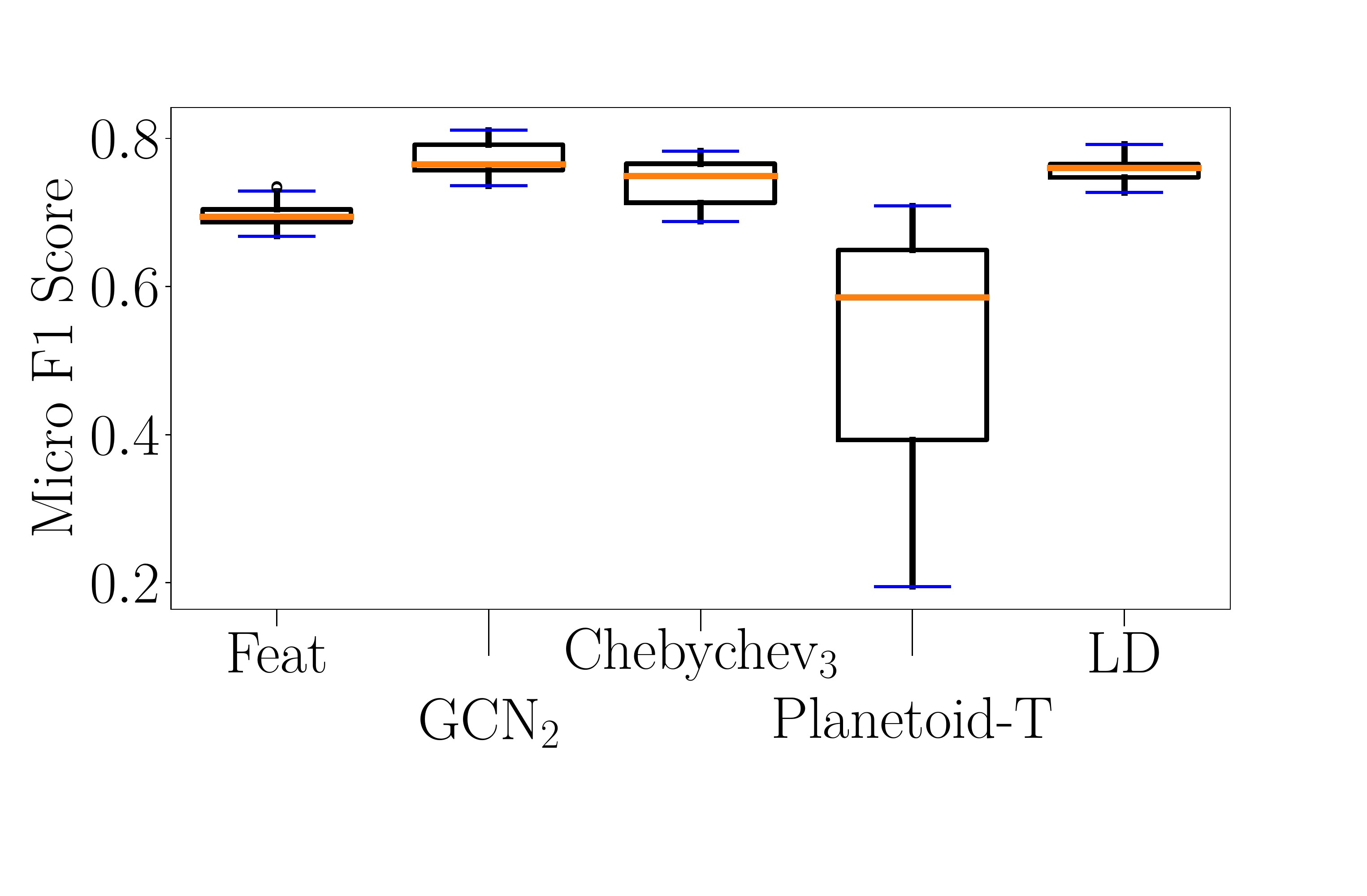}
        \vspace{-7ex}
        \caption{Micro F$_1$ scores for \textit{Pubmed}.}
        \label{fig:pubmed_feat}
    \end{subfigure}
    \caption{Comparison against attribute-based methods: micro F$_1$ scores for the three benchmark data sets.}\label{fig:acc_feat}
\end{figure*}
\subsubsection{Results}
The results for the \textit{BlogCatalog} graph are shown in Figure~\ref{fig:blogcatalog}. For this network, only using the label information from the direct neighborhood of a node is not useful to infer its labels, c.f., \textit{GCN$_1$\_only\_L}. However, incorporating the label distribution of somewhat larger neighborhoods as for our model (again, we also use the APPR matrix calculated for small values of $\alpha$ to determine the label distribution in neighborhoods that span more than 1-hop neighbors) seems to improve the results for the prediciton task significantly. In fact, our model achieves similar, but slightly worse performance than \textit{node2vec} and \textit{DeepWalk}. Given these results, we also combined the node embeddings based on local label distributions with embeddings that capture structural properties. To capture the structural properties we select a very simple approach: we multiply an embedding matrix with the preprocessed adjacency matrix as in Kipf et al. \cite{kipf2016semi}. The embedding matrix is randomly initialized. Note that the structural similarity is defined via direct neighbors. The resulting representation is concatenated with the hidden layer of the \textit{LD} model and the rest of the \textit{LD} model remains the same. The embedding weights are learned jointly with rest of the model.
Having a look at the scores for the resulting model, denoted as \textit{LD+EMB}, this combination further improves the outcome of the prediction.

For the \textit{IMDb Germany} network, for which the results can be seen in Figure~\ref{fig:imdb}
, the labels of even very local neighborhoods are already very expressive. Recalling how this network is constructed, we can expect the latter fact and also the superior performance of our model over the two random walk based methods. Particularly noteworthy for this network is the gain of accuracy that the combination of information from both sources, label distribution and structural properties, achieves.

\subsection{Comparison to Attribute-Based Methods}
To show the power of incorporating label information into the generation process for node embeddings, we also compare our model against the following state-of-the-art attribute-based methods:
\begin{itemize}
\item \textit{Feat}: a baseline approach which predicts node labels only based on the node attributes without considering the underlying graph structure (borrowed from \cite{yang2016revisiting})
\item \textit{GCN$_2$}: the standard 2-layer GCN as published in \cite{kipf2016semi}
\item \textit{Chebychev$_3$}: the spectral convolution method which uses chebychev filters as presented in \cite{NIPS2016_6081}; as in \cite{kipf2016semi} we also use 3rd order chebychev filters
\item \textit{Planetoid-T}: the semi-supervised Planetoid framework which uses attribute information as proposed in \cite{yang2016revisiting}
\end{itemize}
For this set of experiments, we again perform multiclass prediciton on the three benchmark text classification datasets and report the prediction accuracy in terms of micro F$_1$ scores to measure the quality of the retrieved node representations. Note that in contrast to the competitors, our model still does not make use of the node attribute information.
The results are depicted in Figure~\ref{fig:acc_feat} and clearly show that our model can definitely compete with the attribute-based methods and hence is a powerful alternative in cases when no node attributes are present.

\subsection{Impact of the $\alpha$ Parameter}
\begin{figure*}[htb!]
    \centering
    \begin{subfigure}[t]{0.45\columnwidth}
        \includegraphics[width=\textwidth]{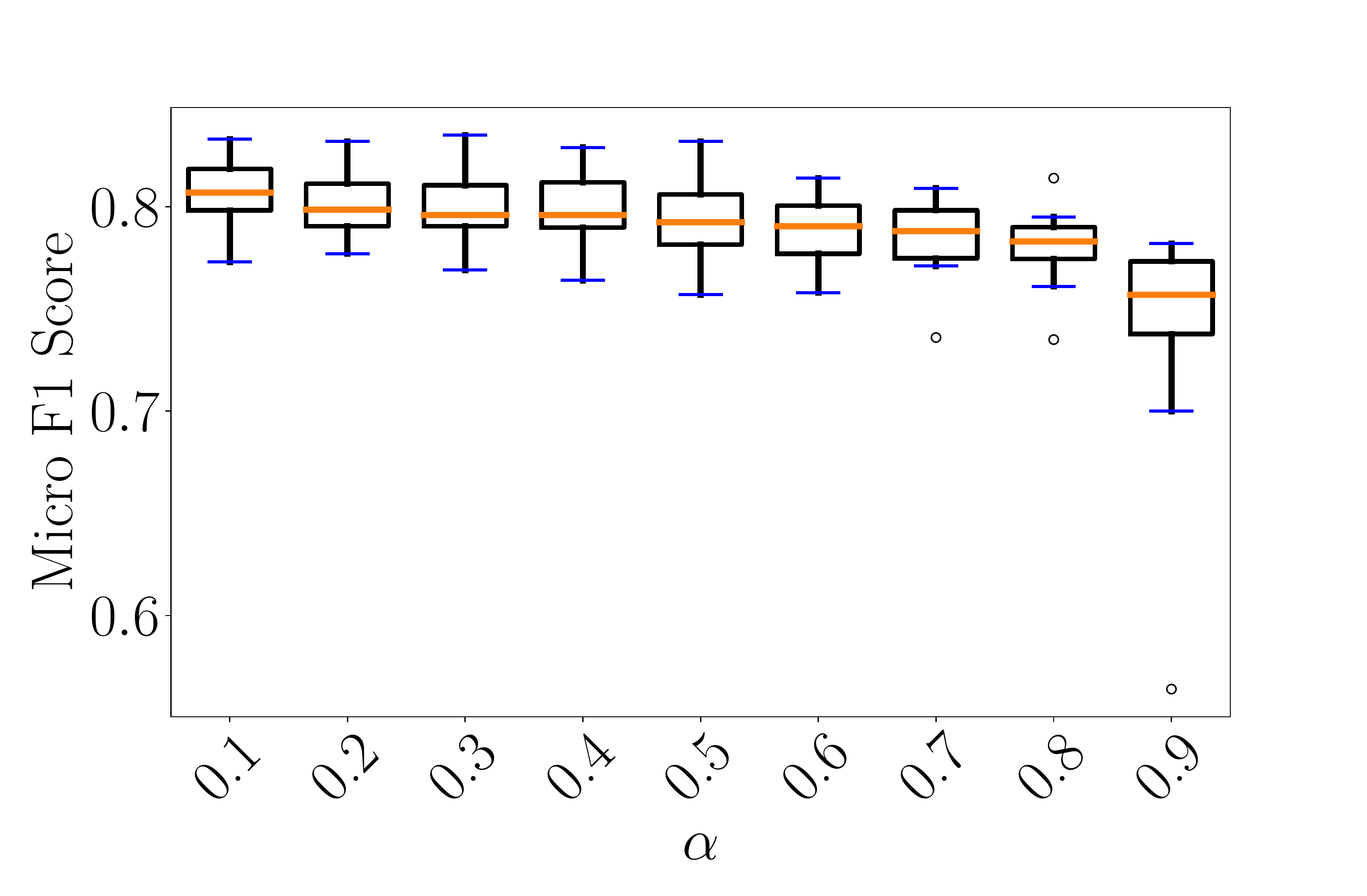}
        \vspace{-5ex}
        \caption{Micro F$_1$ scores for \textit{Cora}.}
        \label{fig:cora_feat}
    \end{subfigure}
    \begin{subfigure}[t]{0.45\columnwidth}
        \includegraphics[width=\textwidth]{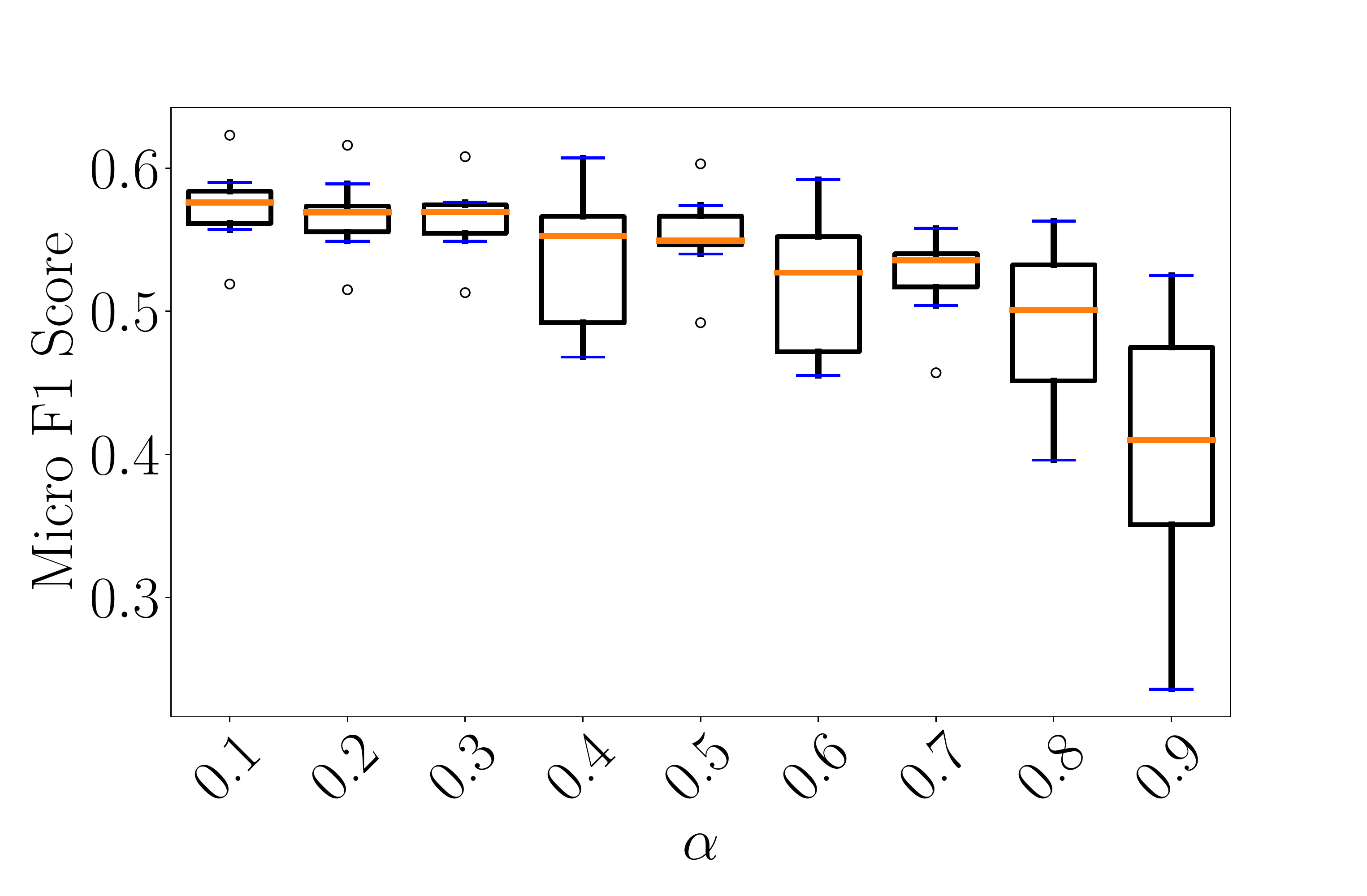}
        \vspace{-5ex}
        \caption{Micro F$_1$ scores for \textit{CiteSeer}.}
        \label{fig:citeseer_feat}
    \end{subfigure}
    \begin{subfigure}[t]{0.45\columnwidth}
        \includegraphics[width=\textwidth]{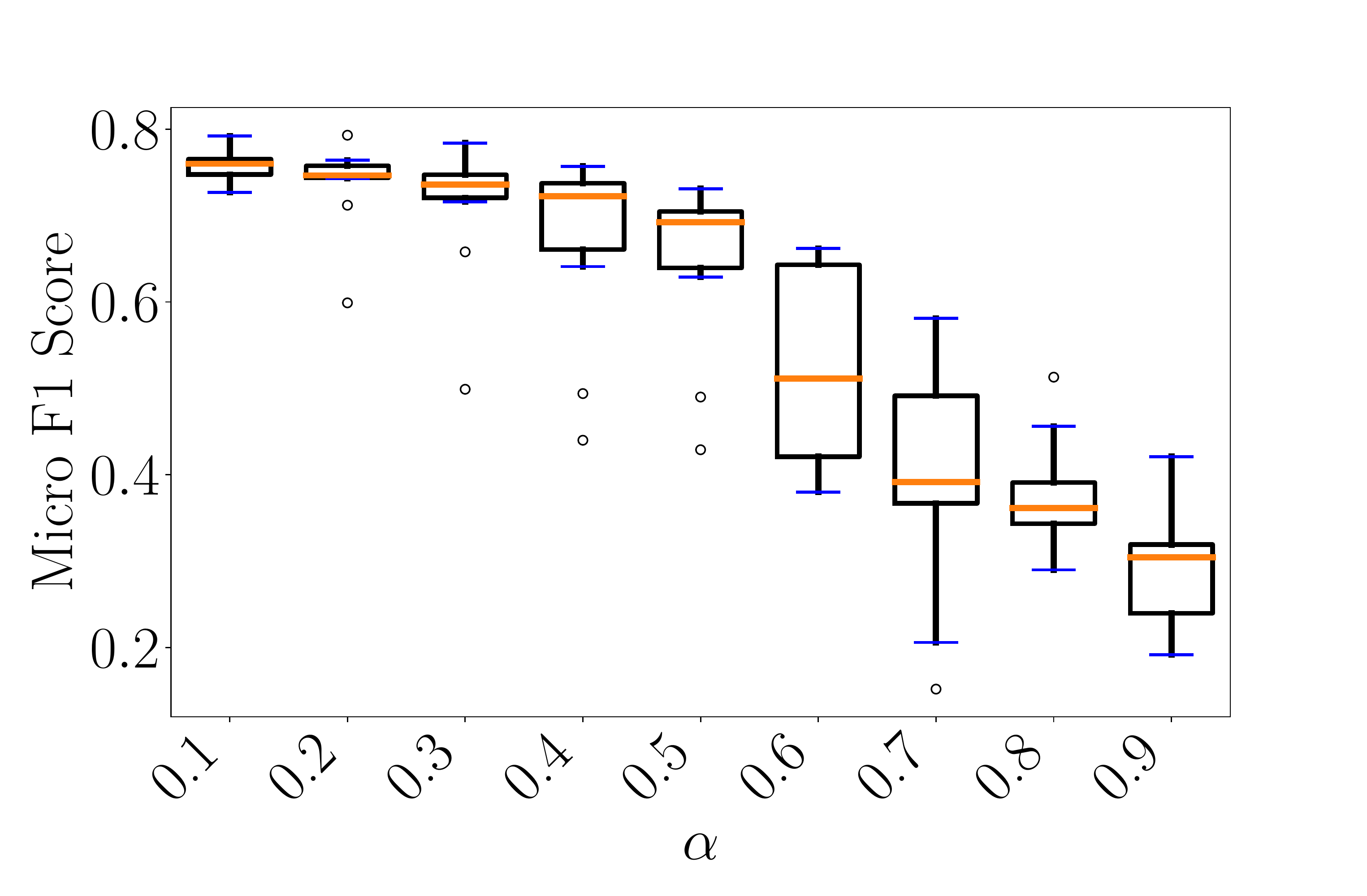}
        \vspace{-5ex}
        \caption{Micro F$_1$ scores for \textit{Pubmed}.}
        \label{fig:pubmed_feat}
    \end{subfigure}
    \caption{Micro F$_1$ scores for the three benchmark data sets when considering different locality levels for node neighborhoods.}\label{fig:diff_alphas}
\end{figure*}

Figure~\ref{fig:diff_alphas} depicts the micro F$_1$ scores achieved for different values of the teleportation parameter $\alpha$ on the three benchmark datasets. As can be seen, particularly for the \textit{Pubmed} network, the model is quite sensitive to the choice of this parameter. Recall that the teleportation parameter determines how far the neighborhood of each node shall be taken into consideration to get the label distributions for each node. Therefore it might make sense to set the $\alpha$ parameter to a small value so that more labels are collected which in turn leads to a more accurate estimation of the local label distribution. On the other hand, this may not hold in every scenario, for instance if the distribution of classes is heterogeneous, i.e., some classes may only appear in areas of the graph where classes are concentrated locally, while other classes may appear in areas where many classes are mixed even within local neighborhoods. An interesting direction for future work is therefore to optimize for some ``good'' $\alpha$ value in a data-driven manner. This may be done either by pre-defining a set of different values of $\alpha$ and approaching for the best of these, or by trying to optimize for some ``good'' $\alpha$ value during the learning procedure. Also, the underlying task, e.g., node classification, may benefit from finding ``good'' values of $\alpha$ for each node individually rather than relying on a global solution.

%% file: conclusion.tex
\vspace{-1ex}
\section{Conclusion}
\label{sec:conclusion}

In this paper, we have introduced a novel label-based approach for semi-supervised node classification. In particular, our method aggregates labels from local neighborhoods using APPR. Most existing approaches consider nodes to be similar, if they are closely related in the graph. Methods for attributed graphs additionally take attributes of the neighboring nodes into account. In contrast, our method can relate nodes even if they are not close-by in the graph and makes more effective use of the labels provided for training to improve the classification quality for graphs with and without node attributes. It is further applicable to nodes unseen during training. The results of our experiments on various real-work datasets demonstrate that local label distributions are able to significantly improve classification results in the multiclass and multilabel setting. Our model is even competitive with state-of-the-art models, which take node attributes into consideration. In a first experiment on multilabel datasets, we were already able to significantly boost the performance by using a simple combination of our model with node embeddings.

For future work, we plan to address the problem of selecting a suitable teleportation parameter $\alpha$.
The $\alpha$ parameter controls the extend of the considered local neighborhood and often has a significant impact on the prediction quality. Performing a grid search to determine a good parameter value is a time consuming task. Furthermore, for different classes varying teleportation parameters might yield the best results.
 
We also aim at further improving the prediction accuracy by further investigating how to effectively combine label-based features with different other kinds of other features, such as node attributes, edge attributes or node embeddings in a semi-supervised model.
Our approach could also be extended to solve additional graph learning tasks, such as link prediction or identification of nodes with unexpected labels for detecting labeling errors or outlier nodes.

\vspace{-1ex}